\documentclass[conference]{IEEEtran}

\usepackage{amsmath,algorithm,algorithmic,amsmath,color,comment,courier,dcolumn,float,graphicx,helvet,latexsym,subfigure,tikz,times,url}

\def\BibTeX{{\rm B\kern-.05em{\sc i\kern-.025em b}\kern-.08em
    T\kern-.1667em\lower.7ex\hbox{E}\kern-.125emX}}

\title{EmTaggeR: A Word Embedding Based Novel Method for Hashtag Recommendation on Twitter}
\author{\IEEEauthorblockN{
Kuntal Dey\IEEEauthorrefmark{1}\IEEEauthorrefmark{2}, 
Ritvik Shrivastava\IEEEauthorrefmark{3},
Saroj Kaushik\IEEEauthorrefmark{2} and
L Venkata Subramaniam\IEEEauthorrefmark{2}}
\IEEEauthorblockA{\IEEEauthorrefmark{1}IBM Research India, New Delhi, India. Email: \{kuntadey,lvsubram\}@in.ibm.com}
\IEEEauthorblockA{\IEEEauthorrefmark{2}Indian Institute of Technology, New Delhi, India. Email \{anz138579,saroj\}@cse.iitd.ac.in}
\IEEEauthorblockA{\IEEEauthorrefmark{3}Netaji Subhas Institute of Technology, New Delhi, India. Email: ritviks.it@nsit.net.in}
}

\date{}

\begin{document}
\maketitle

\begin{abstract}
The hashtag recommendation problem addresses recommending (suggesting) one or more hashtags to explicitly tag a post made on a given social network platform, based upon the content and context of the post.
In this work, we propose a novel methodology for hashtag recommendation for microblog posts, specifically Twitter.
The methodology, EmTaggeR, is built upon a training-testing framework that builds on the top of the concept of word embedding.
The training phase comprises of learning word vectors associated with each hashtag, and deriving a word embedding for each hashtag.
We provide two training procedures, one in which each hashtag is trained with a separate word embedding model applicable in the context of that hashtag, and another in which each hashtag obtains its embedding from a global context.
The testing phase constitutes computing the average word embedding of the test post, and finding the similarity of this embedding with the known embeddings of the hashtags.
The tweets that contain the most-similar hashtag are extracted, and all the hashtags that appear in these tweets are ranked in terms of embedding similarity scores.
The top-$K$ hashtags that appear in this ranked list, are recommended for the given test post.
Our system produces F1 score of 50.83\%, improving over the LDA baseline by around 6.53 times, outperforming the best-performing system known in the literature that provides a lift of 6.42 times.
EmTaggeR is a fast, scalable and lightweight system, which makes it practical to deploy in real-life applications.
\end{abstract}

\section{Introduction}
\label{sec:intro}

\subsection{Background and Motivation}
\label{subsec:motivation}

% Overview
The hashtag recommendation problem addresses recommending (suggesting) one or more hashtags to explicitly tag a post made on a given social network platform, based upon the content and context of the post.
Practically, the hashtags that users tend to assign to a given social network post, depend upon their perception of one or more key facets of the message content.
A large number of social posts, such as tweets, do not contain hashtags, as observed by Hong {\it et al.} \cite{hong2011language}.
And yet, hashtags provide significant value addition, and act as one of the fundamental information facets associated with user-generated social network messages.
As noted in the literature, problems such as topic modeling \cite{asur2011trends}, information diffusion \cite{tsur2012s} \cite{starbird2012will} and many other problems as observed by the literature survey conducted by Dey {\it et al.} \cite{dey2017literature}, can be solved well by using the information content and context of usage of hashtags.
An interesting observation made by \cite{zangerle2011recommending} is that, recommending hashtags ``aim at encouraging the user to (i) use hashtags at all, (ii) use more appropriate hashtags and (iii) avoid the usage of synonymous hashtags''.
All of these indicate that the hashtag recommendation problem is important.

% Introduction to the literature
The problem of hashtag recommendation has emerged as a mainstream area of research over time.
Works such as \cite{zangerle2011recommending}, \cite{huang2012automatic} and \cite{ding2013learning} mark some of the early research efforts in this space.
Later, other methods emerged, such as tweet content hyperlink based ones \cite{sedhai2014hashtag}, Dirichlet-based ones \cite{gong2015hashtag} and topic-based ones \cite{she2014tomoha} \cite{zhang2014time}.
Deep learning based works also started appearing in the literature, such as the work by Weston {\it et al.} \cite{weston2014tagspace}.
The recent deep learning based work by Gong and Zhang \cite{gong2016hashtag}, with a convolutional neural network (CNN), has produced the best-konwn results in the literature.

\subsection{Our Proposition}
\label{subsec:proposition}

% Enabler of our approach
While Dirichlet based approaches that consider topic models, as well as deep CNN based learning approaches, exist in the literature, we observe that the textual context of the words (which in turn constitute the body of the tweets) leave some scope for exploration.
Specifically, word embedding, a recent approach that was proposed by Bengio {\it et al.} \cite{bengio2003neural}, provides a dense, low-dimensional vector based approach, which is effective in storing contextual information within this low-dimensional vector.
The emergence of Word2Vec by Mikolov {\it et al.} \cite{mikolov2013efficient} has enabled the unsupervised word embedding-based approach to be widely adopted by several bodies of research, solving different problems.
We hypothesize that applying word embedding to better understand (learn) the context of hashtag usage behavior (with respect to the words used along with the hashtag), and using this learning to recommend hashtags for test tweets, is likely to be an effective approach.
Although prior works on hashtag recommendation, namely Weston {\it et al.} \cite{weston2014tagspace} and Gong and Zhang \cite{gong2016hashtag}, use embedding as part of their overall solution design, we use embedding in a different and novel manner that integrates more deeply with the user-generated content, by deriving the embeddings of hashtags using the content words appearing with the hashtag.

% The core of our approach
We propose a model to train for learning word embeddings and testing (assigning hashtags) with the trained embedding model, for the task of recommending hashtags towards tweets.
We learn word embeddings from the given traiing data in the training phase.
We develop two different training models in the current work.
In one model, we globally train the embedding for the words that appear across the entire vocabulary.
In the other model, we create corpus for each specific hashtag, and learn the embedding of each word within the scope of that hashtag.
We use the skip-gram model for training the embeddings.

In the testing phase, the embeddings are extracted from the trained model, for all words associated with the test tweet.
The ``embedding'' of the test tweet is computed as the average vector of the extracted embeddings.
The similarity of these embeddings are compared with the embeddings of the hashtags learned during the training phase, and the tweets containing the best-matching hashtag are extracted.
The hashtags of these tweets are collected, and ranked using the embedding similarity that was already computed earlier in the testing process.
The top $K$ hashtags with the highest similarity measures are recommended as the hashtags for the given test tweet.
For the first model, the word embedding vectors are chosen from the global occurrences of the word.
For the second model, the word embedding vectors are chosen from the occurrences of the word that are local with respect to the hashtag under consideration for recommendation.

We perform empirical validation of our approach, using real-life Twitter data.
In absence of any benchmark data in this space, we assess the goodness of our system, and compare it with the literature, using {\it performance lift values} over the well-adopted Latent Dirichlet Allocation (LDA) \cite{blei2003latent} baseline.
Performance lift of a method, captures the ratio of performance given by that method (such as, our method EmTaggeR) to the performance given by a baseline method (the LDA baseline, in this case).
While the state of the art \cite{gong2016hashtag} provides a lift of $6.42$ times over a LDA baseline, our approach provides a lift of $6.53$ times.
Thus, our work establishes a new benchmark.

\section{Related Work}
\label{sec:relwork}

% Early works - inception
Hashtag recommendation has been a key research problem for long.
An early work by Davidov {\it et al.} \cite{davidov2010enhanced}, that focused around sentiment classification, had touched upon the angle of hashtags with sentiments.
One of the first works that focused completely on hashtag prediction, was carried out by Zangerle {\it et al.} \cite{zangerle2011recommending}.
The work attempts to recommend hashtags by (a) computing a {\it tf-idf} based content similarity of a target tweet with other tweets that exist in the database, (b) retrieving the tweets of the most similar messages and (c) ranking and recommending the hashtags that appear within these tweets.
Ding {\it et al.} \cite{huang2012automatic} \cite{ding2013learning} propose an unsupervised learning method using a latent variable estimation based topical translation model.
They model by treating hashtags and tweet content as a parallel occurrence of a target concept.

Sedhai and Sun \cite{sedhai2014hashtag} recommend hashtags for tweets, that contain hyperlinks as part of their content.
They propose a two-phase solution.
In the first phase, they select a set of candidate hashtags using several attributes, such as consideration of similar tweets, hyperlinked documents, named entities and the domain of the content of the webpage that the hyperlink refers to.
In the second phase, they formulate as a learning-to-rank problem, and solve with RankSVM to aggregate and rank the candidate hashtags selected in the first phase.

Gong {\it et al.} \cite{gong2015hashtag} proposed a Dirichlet based mixture model, incorporating types of hashtags as hidden variables.
They propose their framework as a non-parametric Bayesian method, and motivated by Liu {\it et al.} \cite{liu2012topical} they base their model under the assumption that hashtags and tweet content are parallel descriptions of the same content.
This is also similar to the philosophy of \cite{huang2012automatic} and \cite{ding2013learning}.

Works such as \cite{she2014tomoha} and \cite{zhang2014time} propose topic-based hashtag recommendation models.
She and Chen \cite{she2014tomoha} perform supervised topic model learning using the hashtags as topic labels, to discover inter-word relationships.
They infer the probability that a hashtag will be contained in a new tweet, by treating words either as background words (that are prevalent in many tweets) or local topic words (that are more specific to that tweet), and generate hashtags for recommendation using a symmetric Dirichlet distribution of the local and background words.
Zhang {\it et al.} \cite{zhang2014time} extend upon the widely used translation model for hashtag recommendation in the literature, to include the aspects of temporal and personal factors.
They draw from a multinomial word-topic distribution and retain (recommend) the hashtags with the maximum probabilities.

The concept of word embedding was introduced by Bengio {\it et al.} \cite{bengio2003neural}.
The concept was enriched by the work of Mikolov {\it et al.} \cite{mikolov2013efficient}, where they proposed the now-popular Word2Vec model.
This has enabled the unsupervised word embedding-based approach to be widely adopted by several bodies of research, solving different problems.
The concept of embedding also acts as the foundation of the hashtag recommendation works by Weston {\it et al.} \cite{weston2014tagspace} and Gong and Zhang \cite{gong2016hashtag}.
Amongst other related approaches, GloVe \cite{pennington2014glove} based embedding has also found research traction.

Deep neural network based approaches have also emerged in the literature.
Weston {\it et al.} \cite{weston2014tagspace} model a deep CNN based architecture that learns semantic embeddings from hashtags.
Their model represents the words, as well as the entire textual posts, as embeddings in the intermediate layers of their deep-CNN architecture.
The philosophy of this work is similar to ours; however, there are significant differences such as, we use a per-hashtag word embedding, as well as, we use skip-gram embedding instead of a bag-of-words based one - all of these factors play a major role in improving our performance.
In a recent work that has established the current state of the art, Gong and Zhang \cite{gong2016hashtag} proposed the use of a deep CNN with embedding and attention mapping.

While the recent works that use embedding, tend to perform other intelligent actions such as adding the entire textual post also along with the embedding \cite{weston2014tagspace} and local attention maps \cite{gong2016hashtag}, our approach is different in having hashtag-specific word embeddings.
These works compute the embeddings of hashtags as a direct derivation of the word neighborhoods of the hashtags, as opposed to our approach where we use the constituent words' embeddings to derive embeddings of hashtags.
Our approach adds context, by proposing a per-hashtag word embedding structure in one of the models, making it unique and novel, and outperforming the state of the art.
We do not perform any convolution operation, and rely upon the inherent recurrent neural network (RNN) based contextualization that are used to find embeddings.

\section{Our Approach}
\label{sec:algo}

\begin{figure*}[htb]
\centering
\includegraphics[width=0.9\textwidth]{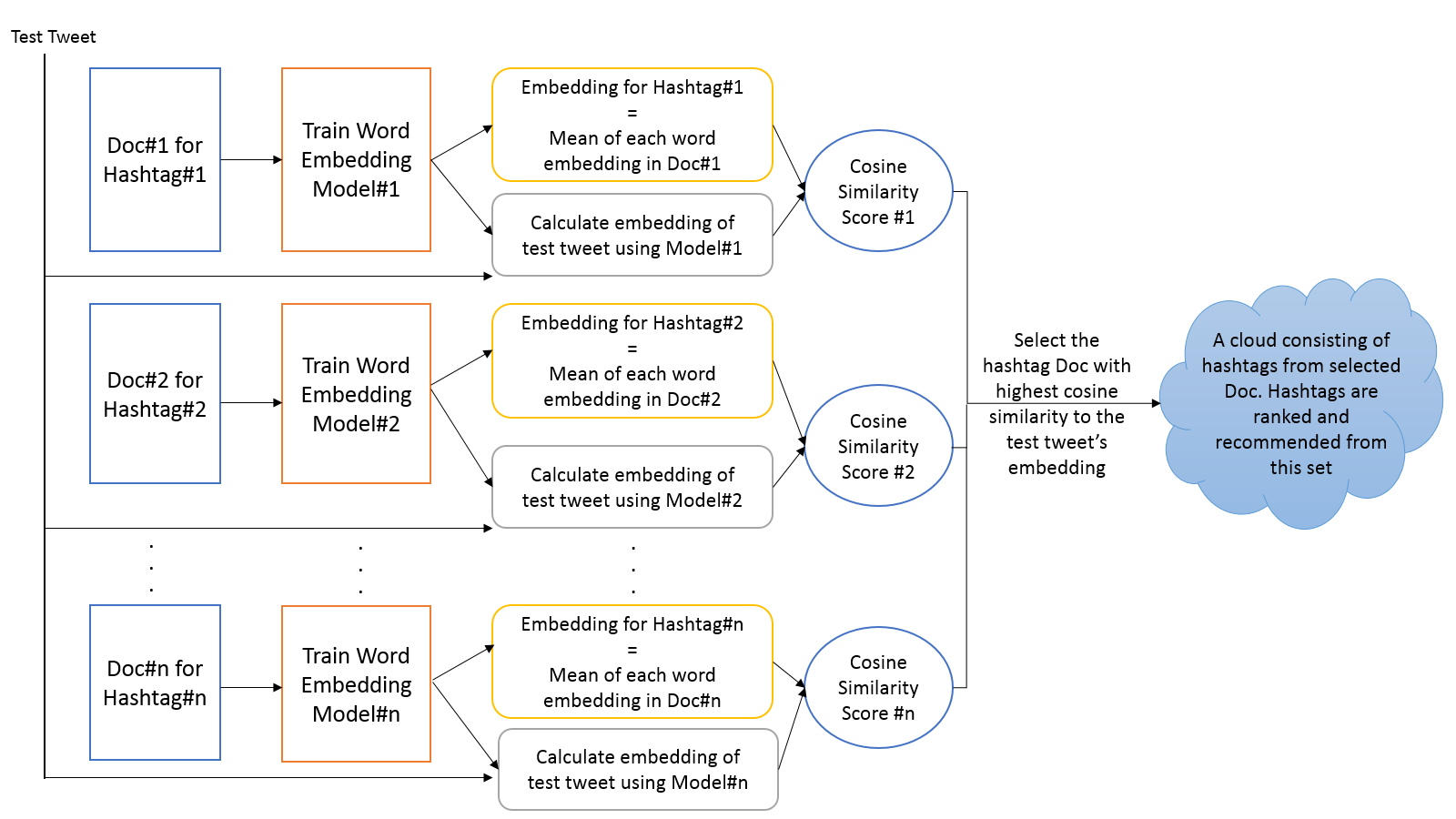}
\vspace{-0.1in}
\caption{System Architecture Diagram with Model 1}
\label{fig:model1}
\vspace{-0.15in}
\end{figure*}

\begin{figure*}[htb]
\centering
\includegraphics[width=0.9\textwidth]{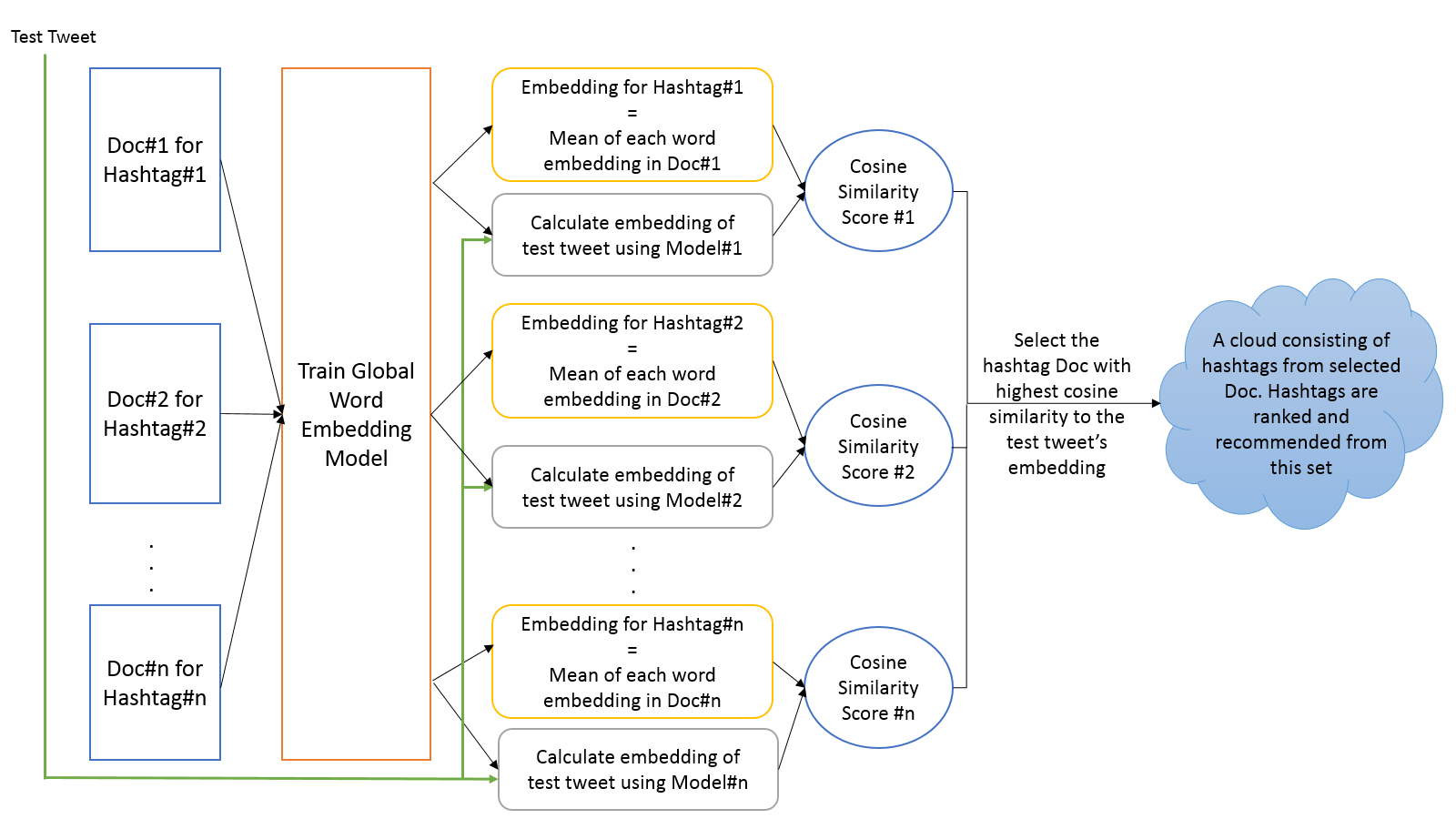}
\vspace{-0.1in}
\caption{System Architecture Diagram with Model 2}
\label{fig:model2}
\vspace{-0.15in}
\end{figure*}

We present the details of our approach in this section.
After initial preprocessing of the input microblog (Twitter) data, we perform a train-test based approach.
Overall, the training is performed over two phases: (a) learning the embedding of each word, and, (b) learning the embedding of each hashtag, using the embedding of the constituent words.
The embedding of each word is learned using two different methods - one from a global corpus that is constructed as a union of all the available tweets, and another from a hashtag-specific usage of words.
Below, we provide an introduction to word embedding, which is already present in the literature and is at the core of our model.
We subsequently provide the details of our approach.

\subsection{Data Cleaning and Preprocessing}
\label{subsec:cleanpreproc}

\subsubsection{Data Cleaning}
\label{subsub:cleaning}

We first perform data cleaning, in order to make the provided tweets useful for our work.

\begin{itemize}
\item \textbf{Non-English tweet removal:} We eliminate the non-English tweets from our dataset. We detect non-English tweets using the language marker meta-data, that is an integral part of the raw Twitter data.
\item \textbf{Non-ASCII character removal:} We eliminate the non-ASCII content that is present in any of the English tweets (retained in the earlier step).
\item \textbf{Removal of tweets without any hashtag:} Since the objective of our work, namely hashtag recommendation, will need tweets that have at least one hashtag for training, and the hashtags recommended for the test tweets will need to be compared with a ground truth (real hashtags that were given by the users to those tweets), it is necessary to retain only those tweets that contain at least one hashtag. Hence, we eliminate each tweet that does not have at least one hashtag associated with it.
\end{itemize}

\subsubsection{Preprocessing}
\label{subsub:preprocessing}

We perform text stemming, tweet normalization (including noise removal to an extent), and stopword removal, as part of the preprocessing phase.

\begin{itemize}
\item \textbf{Stemming:} We stem the tokens present in the tweet content. Since, hashtags typically do not tend to be characterized by what stemming eliminates (such as, tenses, numbers {\it etc.}), stemming does not lose the necessary information in the context of hashtag recommendation.
\item \textbf{Tweet normalization:} We perform normalization of the tweet content, by resolving many colloquial on-the-net expressions that appear on user-generated social media text, but do not appear in any traditional dictionary. For instance, what appears as {\it aaf} on Twitter, is expanded to {\it as a friend}. We use an online net slang resolution dictionary\footnote{http://www.noslang.com/dictionary}, along with Han-Baldwin dictionary \cite{han2011lexical}, for tweet normalization.
\item \textbf{Stopword removal:} We use an online resource from the Stanford NLP resources for stopword removal\footnote{https://nlp.stanford.edu/IR-book/html/htmledition/dropping-common-terms-stop-words-1.html}.
\end{itemize}

The content generated as the output of preprocessing, is sent to the next phase of the system, to learn embeddings.
We propose two different (but similar) models for embedding.

\subsection{Model 1: Per Hashtag Word Embedding Based}
\label{subsec:model1}

In this model, we create a per-hashtag word embedding matrix, where, each word appearing along with a hashtag, has a different embedding with respect to that hashtag.
We derive the embedding of a hashtag, as the average of all hashtag-specific embeddings of all the words appearing for that hashtag.
The architecture of the figure is illustrated in Figure~\ref{fig:model1}.
The training process is carried out over the following steps.

\subsubsection{Training - Document Creation}
\label{subsubsec:model1-document}

A document is created for each hashtag, by first collecting all the tweets $t_h$ where the given hashtag $h$ appears, and then appending all the tweets thus obtained.
Note that, all the hashtags and mentions are eliminated from the document.
Let $H$ be the set of all the hashtags that appear in the overall dataset.
Thus for each hashtag $h$, we create a document $_1D_h$ as
\begin{equation}
_1D_h = \mathop{\bigcup}\{t_h\} - ({\forall h \in H})\{h\}
\label{eqn:1Dh}
\end{equation}

\subsubsection{Training - Word and Hashtag Embedding Determination}
\label{subsubsec:model1-embedding}

Next, a word embedding model is created for each document (corresponding to a hashtag).
To create this, each word is assigned an embedding based upon its usage across only those tweets that contain this hashtag, {\it i.e.}, belong to the document for this hashtag.
We perform skip-gram based Word2Vec embedding \cite{mikolov2013efficient}, and do not use the bag-of-word embeddings model at all.
If $E$ denotes the embedding learning operation on a word from a given document $_1D_h$, $_1v_{w,h}$ is the embedding vector of a word $w$ in the context of hashtag $h$, $s$ is the skip window size of the skip-gram, $c_w$ is the minimum occurrence count of a word in the document $_1D_h$ for being considered at all, and $l_v$ is the dimension of the embedding vector that will be derived at the output, then the embedding of the word in the hashtag (the corresponding document) is computed as:
\begin{equation}
_1v_{w,h} = E(_1D_h, s, c_w, l_v)
\label{eqn:1Ew}
\end{equation}

Let $W_h$ be the set of words appearing in $_1D_h$.
The embedding computation operation is repeated for all words $w \in W_h$, to obtain the embedding $_1v_{w,h}$ of all the words $w$ that appear within the hashtag.
Finally, an embedding $_1V_h$ is computed for each given hashtag as a whole, using the embedding of the words that appear in the tweets containing the hashtag.
This is simply computed as the average of all the word embeddings appearing in its document.
\begin{equation}
_1V_h = \frac{\sum\limits_{w \in _1D_h}^{}(_1v_{w,h})}{|_1D_h|}
\label{eqn:1Eh}
\end{equation}

In the above, $|_1D_h|$ denotes the total length of the document $_1D_h$ in terms of number of words (with repeated words retained).
The above is repeated for all hashtags $h \in H$, thus creating a complete embedding map, for all the words $w$ under the context of all the hashtags $h$ that they appear in.
Note that, words that are repeated in the document, are considered as many times as they appear; that is, the repeating behavior is retained, as this inherently assigns the necessary weight that the embedding merits with respect to how highly each word is used with respect to that hashtag.
The training process for Model 1 has been summarized in Algorithm~\ref{alg:emtaggertraina}.

\algsetup{indent=2em}
\newcommand{\emtaggertraina}{\ensuremath{\mbox{\sc EmTaggeR: Model 1 Training}}}
\begin{algorithm}[t]
\begin{algorithmic}[1]
\STATE {\it function} {\bf EmTaggeRTrainModel1():}
\FOR {hashtag $h$ in $H$}
	\STATE $_1D'_h \gets$ Append all tweets where hashtag $h$ appears
	\STATE $_1D_h \gets _1D'_h - H$ (H is the set of all hashtags)
	\STATE $s \gets$ Skip window size
	\STATE $c_w \gets$ Min. required word occurrence count in $_1D_h$
	\STATE $l_v \gets$ Desired Word2Vec length
	\STATE Embedding of word $w$ in hashtag $h$: $_1v_{w,h} \gets E(_1D_h, s, c_w, l_v)$
	\STATE Embedding of hashtag $h$: $_1V_h \gets$ Average embedding $\frac{\sum\limits_{w \in _1D_h}^{}(_1v_{w,h})}{|_1D_h|}$ of all words $w$
\ENDFOR
\renewcommand{\algorithmicreturn}{\textbf{Output:}}
\RETURN $(\forall{h \in H}) _1V_h$
\caption{$\emtaggertraina$}
\label{alg:emtaggertraina}
\end{algorithmic}
\end{algorithm}

\subsubsection{Testing the Model}
\label{subsubsec:model1-testing}

The aim of the testing (hashtag recommendation) phase is to align each test tweet to the best possible hashtag(s), and assign the most highly aligned hashtag(s) to the test tweet.
We compute the alignment of each given hashtag to a given test tweet as follows.
Given a test hashtag $h$, a set of words $W_h$ associated with hashtag $h$ that have their embeddings as part of $_1v_{w,h}$, for each word $w_t$ in the test tweet $t_t$, we retrieve the embedding of $w_t$ in the word embedding $_1v_{w,h}$ trained for that hashtag.
That is, we retrieve $_1v_{w_t,h}$, an embedding for word $w_t$ trained with respect to hashtag $h$, as:
\begin{equation}
_1v_{w_t,h} = _1v_{w,h},\forall{w_t \in W_h}
\label{eqn:1Ewt}
\end{equation}

The embedding of the test tweet is derived by computing the average of all the word embeddings appearing in the test tweet, that are also present in the training vocabulary.

\begin{equation}
_1V_{h_t} = \frac{\sum\limits_{w_t \in t_t}^{}(_1v_{w_t,h})}{|_1v_{w_t,h}|}
\label{eqn:1Eht}
\end{equation}

We compute the similarity of the hashtag's embedding with the test tweet's embedding within that hashtag, and select the hashtag that have the highest value.
If $\sigma(h_t, h)$ denotes the similarity (such as, cosine similarity) of the embedding of a test hashtag $h_t$ with that of a given hashtag $h$, then, the hashtag similarity scores with respect to the test tweet, are obtained as
\begin{equation}
h_{sim} = \forall{h_t \in H} \sigma(h, _1V_{h_t})
\label{eqn:1sim}
\end{equation}

The best-matching hashtag is selected as:
\begin{equation}
g = \forall{h_t \in H} max(h_{sim})
\label{eqn:1max}
\end{equation}

In one embodiment of the problem where we recommend the single-best hashtag, we terminate our testing process here, and output $g$.
However, in another embodiment where we need to recommend top-$K$ hashtags, we perform the following.
We go back to the training tweets, and pick all the training tweets that contain the hashtag $g$, and form a set union $g_t$ of all the hashtags that these hashtags together contain.
We rank the hashtags based upon the $h_{sim}$ list, and pick $G$, the top-$K$ hashtags that appear also in $g_t$.
$G$ is the set of hashtags provided as the output of Model 1.
The testing process using Model 1 has been summarized in Algorithm~\ref{alg:emtaggertesta}.

\algsetup{indent=2em}
\newcommand{\emtaggertesta}{\ensuremath{\mbox{\sc EmTaggeR: Model 1 Testing}}}
\begin{algorithm}[t]
\begin{algorithmic}[1]
\STATE {\it function} {\bf EmTaggeRTestModel1():}
\STATE MaxAlignedHashtag $\gets \phi$
\STATE MaxAlignmentScore $\gets 0$
\STATE Let $t_t$ be a test tweet, $w_t$ the words in $t_t$
\FOR {hashtag $h \in H$}
	\STATE $W_h \gets$ Vocabulary (set of words) across all tweets that contain $h$
	\FOR {$\forall{w_t} \in t_t$}
		\STATE Retrieve embedding $_1v_{w_t,h}$ for word $w_t$ in $h$: $_1v_{w_t,h} = _1v_{w,h}, (\forall{w_t \in W_h})$
		\STATE Find average embedding of test tweet: $_1V_{h_t} \gets \frac{\sum\limits_{w_t \in t_t}^{}(_1v_{w_t,h})}{|_1v_{w_t,h}|}$
		\STATE Compute similarity of $t_t$ and $h_t$: $h_{sim} \gets \sigma(h, _1V_{h_t})$
		\IF {$h_{sim} >$ MaxAlignmentScore}
			\STATE MaxAlignedHashtag $\gets h_t$
			\STATE MaxAlignmentScore $\gets h_{sim}$
		\ENDIF
	\ENDFOR
\ENDFOR
\STATE $g \gets$ MaxAlignedHashtag
\IF {only one hashtag needs to be recommended}
	\STATE $G \gets g$
\ELSE
	\STATE Pick all training tweets having hashtag $g$
	\STATE Sort (rank) $h_{sim}$
	\STATE $G$ $\gets$ The top-$K$ hashtags as per $h_{sim}$ values that appear also in $g_t$
\ENDIF
\renewcommand{\algorithmicreturn}{\textbf{Output:}}
\RETURN $(\forall{h \in H}) _1V_h$
\caption{$\emtaggertesta$}
\label{alg:emtaggertesta}
\end{algorithmic}
\end{algorithm}

\subsection{Model 2: Cross Hashtag Word Embedding Model}
\label{subsec:model2}

In this model, we create a global word embedding matrix, where, each word has an embedding agnostic to hashtags.
We derive the embedding of a hashtag, as the average of all global embeddings of all the words appearing for that hashtag.
The architecture of the figure is illustrated in Figure~\ref{fig:model2}.
The training for this model is performed as follows.

\subsubsection{Training - Document Creation}
\label{subsubsec:model2-document}

A single global document is constructed by appending all the tweets irrespective of the hashtags they contain.
In this model too, all the hashtags and mentions are eliminated from the document.
Let $H$ be the set of all the hashtags that appear in the overall dataset, $h (\in H)$ be the set of individual hashtags, and let $t$ be the set of the tweets.
The document $_2D_H$ is created as
\begin{equation}
_2D_H = \mathop{\bigcup}\{t\} - ({\forall h \in H})\{h\}
\label{eqn:2DH}
\end{equation}

Further, for each hashtag $h$, we create a document $_2D_h$ as
\begin{equation}
_2D_h = \mathop{\bigcup}\{t_h\} - ({\forall h \in H})\{h\}
\label{eqn:2Dh}
\end{equation}

\subsubsection{Training - Word and Hashtag Embedding Determination}
\label{subsubsec:model2-embedding}

Subsequently, a word embedding model is created across all the hashtags, using the single global document created above.
For this, each word is assigned an embedding based upon its usage in the document.
Akin to Model 1, we perform skip-gram based Word2Vec embedding \cite{mikolov2013efficient}, and do not use the bag-of-word embeddings model at all, in Model 2 (the current model) as well.
If $E$ denotes the embedding learning operation on a word from a given document $_2D_H$, $_2v_w$ is the embedding vector of a word $w$ in the document, $s$ is the skip window size of the skip-gram, $c_w$ is the minimum occurrence count of a word in the document $_2D_H$ for being considered at all, and $l_v$ is the dimension of the embedding vector that will be derived at the output, then the embedding of the word $w$ in the hashtag (the corresponding document) is computed as:
\begin{equation}
_2v_w = E(_2D_H, s, c_w, l_v)
\label{eqn:2Ew}
\end{equation}

Let $W_H$ be the set of words appearing in $_2D_H$.
The embedding computation is performed for all words $w \in W_H$, to obtain the embedding $_2v_w$ of all the words $w$.
Finally, an embedding $_2V_h$ is computed for each given hashtag as a whole, using the embedding of the words appear in the content of the tweets containing the hashtag, as the average of all the embeddings of these words.
Then, the embedding of hashtag $h$ is given as:
\begin{equation}
_2V_h = \frac{\sum\limits_{w \in _2D_H}^{}(_2v_w)}{|_2D_h|}
\label{eqn:2Eh}
\end{equation}

Here, $|_2D_h|$ denotes the total length of the document $_2D_h$ in terms of number of words (with repeated words retained), which is the per-hashtag occurrence of the words $w \in _2D_h$.
The above is repeated for all hashtags $h \in H$, thus creating a complete embedding map, for all the words $w$, across all the tweets (ducoments).
Akin to Model 1, repeated words here too are retained, for the same reasons, for computing the embedding of the hashtags.
The training process for Model 2 has been summarized in Algorithm~\ref{alg:emtaggertrainb}.

\algsetup{indent=2em}
\newcommand{\emtaggertrainb}{\ensuremath{\mbox{\sc EmTaggeR: Model 2 Training}}}
\begin{algorithm}[t]
\begin{algorithmic}[1]
\STATE {\it function} {\bf EmTaggeRTrainModel2():}
\STATE $_2D'_H \gets$ Append all tweets
\STATE $_2D_H \gets _2D'_h - H$ (H is the set of all hashtags)
\STATE $s \gets$ Skip window size
\STATE $c_w \gets$ Min. required word occurrence count in $_2D_h$
\STATE $l_v \gets$ Desired Word2Vec length
\STATE Embedding of word $w$: $_2v_w = E(_2D_H, s, c_w, l_v)$
\FOR {hashtag $h$ in $H$}
	\STATE $_2D'_h \gets$ Append all tweets where hashtag $h$ appears
	\STATE $_2D_h \gets _2D'_h - H$ (H is the set of all hashtags)
	\STATE Embedding of hashtag $h$: $_2V_h \gets$ Average embedding $\frac{\sum\limits_{w \in _2D_H}^{}(_2v_w)}{|_2D_h|}$
\ENDFOR
\renewcommand{\algorithmicreturn}{\textbf{Output:}}
\RETURN $(\forall{h \in H}) _1V_h$
\caption{$\emtaggertrainb$}
\label{alg:emtaggertrainb}
\end{algorithmic}
\end{algorithm}

\subsubsection{Testing the Model}
\label{subsubsec:model2-testing}

The testing (hashtag recommendation) process for Model 2 is similar to that of Model 1, except that, the hashtag embeddings are retrieved from the global embeddings rather than hashtag-specific ones.
Here, the testing is performed as follows.

Given a test hashtag $h$, a set of words $W_h$ associated with hashtag $h$ that have their embeddings as part of $_2v_w$, for each word $w_t$ in the test tweet $t_t$, we retrieve the embedding of $w_t$ in the trained word embedding $_2v_w$.
That is, we retrieve an embedding for $w_t$ in $h$, namely $_2v_{w_t,h}$, as:
\begin{equation}
_2v_{w_t,h} = _2v_w,\forall{w_t \in W_h}
\label{eqn:2Ewt}
\end{equation}

The embedding of the test tweet is derived by computing the average of all the word embeddings appearing in the test tweet, that are also present in the training vocabulary (and thus has an embedding trained for it during the training phase, using the single global document).

\begin{equation}
_2V_{h_t} = \frac{\sum\limits_{w_t \in t_t}^{}(_2v_{w_t,h})}{|_2v_{w_t,h}|}
\label{eqn:2Eht}
\end{equation}

The rest of the testing pipeline is precisely the same for Model 1 and Model 2.
Akin to the testing phase of Model 1, for testing Model 2 also we compute the similarity of the hashtag's embedding with the test tweet's embedding within that hashtag, and select the hashtag that have the highest value.
If $\sigma(h_t, h)$ denotes the similarity (such as, cosine similarity) of the embedding of a test hashtag $h_t$ with that of a given hashtag $h$, then, the hashtag similarity scores with respect to the test tweet, are obtained as
\begin{equation}
h_{sim} = \forall{h_t \in H} \sigma(h, _2V_{h_t})
\label{eqn:2sim}
\end{equation}

The best-matching hashtag is selected as:
\begin{equation}
g = \forall{h_t \in H} max(h_{sim})
\label{eqn:2max}
\end{equation}

Akin to the other model, in one embodiment of the problem where we recommend the single-best hashtag, we terminate our testing process here, and output $g$.
However, in another embodiment where we need to recommend top-$K$ hashtags, we perform the following.
We go back to the training tweets, and pick all the training tweets that contain the hashtag $g$, and form a set union $g_t$ of all the hashtags that these hashtags together contain.
We rank the hashtags based upon the $h_{sim}$ list, and pick $G$, the top-$K$ hashtags that also appear in $g_t$.
Akin to Model 1, $G$ is the set of hashtags provided as the output of Model 2.
The testing process using Model 2 has been summarized in Algorithm~\ref{alg:emtaggertesta}.

\algsetup{indent=2em}
\newcommand{\emtaggertestb}{\ensuremath{\mbox{\sc EmTaggeR: Model 2 Testing}}}
\begin{algorithm}[t]
\begin{algorithmic}[1]
\STATE {\it function} {\bf EmTaggeRTestModel2():}
\STATE MaxAlignedHashtag $\gets \phi$
\STATE MaxAlignmentScore $\gets 0$
\STATE Let $t_t$ be a test tweet, $w_t$ the words in $t_t$
\FOR {hashtag $h \in H$}
	\STATE $W_h \gets$ Vocabulary (set of words) across all tweets that contain $h$
	\FOR {$\forall{w_t} \in t_t$}
		\STATE Retrieve embedding $_2v_{w_t,h}$ for word $w_t$ from the global embedding: $_2v_{w_t,h} = _2v_w, (\forall{w_t \in W_h})$
		\STATE Find average embedding of test tweet: $_2V_{h_t} \gets \frac{\sum\limits_{w_t \in t_t}^{}(_2v_{w_t,h})}{|_2v_{w_t,h}|}$
		\STATE Compute similarity of $t_t$ and $h_t$: $h_{sim} = \forall{h_t \in H} \sigma(h, _2V_{h_t})$
		\IF {$h_{sim} >$ MaxAlignmentScore}
			\STATE MaxAlignedHashtag $\gets h_t$
			\STATE MaxAlignmentScore $\gets h_{sim}$
		\ENDIF
	\ENDFOR
\ENDFOR
\STATE $g \gets$ MaxAlignedHashtag
\IF {only one hashtag needs to be recommended}
	\STATE $G \gets g$
\ELSE
	\STATE Pick all training tweets having hashtag $g$
	\STATE Sort (rank) $h_{sim}$
	\STATE $G$ $\gets$ The top-$K$ hashtags as per $h_{sim}$ values that appear also in $g_t$
\ENDIF
\renewcommand{\algorithmicreturn}{\textbf{Output:}}
\RETURN $(\forall{h \in H}) _1V_h$
\caption{$\emtaggertestb$}
\label{alg:emtaggertestb}
\end{algorithmic}
\end{algorithm}

\section{Experiments}
\label{sec:expt}

\begin{table}[htb]
\begin{center}
\begin{tabular}{|l|c|}
\hline
\textbf{Tweet Selection Criteria} & \textbf{Count} \\
\hline
Total number of tweets & $34,114,982$ \\
\hline
Tweets in English & $13,410,808$ \\
\hline
English Tweets containing at least one hashtag & $2,417,163$ \\
\hline
Hashtag filtering (retain if 200-500 occurrences) & $251,649$ \\
\hline
Training tweet dataset size & $175,000$ \\
\hline
Validation tweet dataset size & $25,000$ \\
\hline
Testing tweet dataset size & $51,649$ \\
\hline
\end{tabular}
\end{center}
\caption{Data description}
\label{tab:datadesc}
\vspace{-0.3in}
\end{table}

\begin{table*}[htb]
\begin{center}
\begin{tabular}{|l|c|c|c|c|c|c|}
\hline
\textbf{Model} & $L = 25$ & $L = 50$ & $L = 100$ & $L = 200$ & $L = 400$ & $L = 600$ \\
\hline
Model 1: At-Least-One Correct (ALOC) & 0.4378 & \textbf{0.4395} & 0.4334 & 0.4270 & 0.4199 & 0.4183 \\
Model 1: Multiple Correct (MuC) & 0.3403 & \textbf{0.3411} & 0.3321 & 0.3264 & 0.3228	& 0.3189 \\
Model 2: At-Least-One Correct (ALOC) & 0.5743 & \textbf{0.5829} & 0.5809 & 0.5780 & 0.5700 & 0.5662 \\
Model 2: Multiple Correct (MuC) & 0.4993 & \textbf{0.5083} & 0.5059 & 0.5003 & 0.4958 & 0.493 \\
\hline
\end{tabular}
\end{center}
\caption{F1-score (performance) of the different models, for the different performance measurement metrics used in the paper}
\label{tab:modelperformances}
\vspace{-0.3in}
\end{table*}

\begin{table}[htb]
\begin{center}
\begin{tabular}{|l|c|}
\hline
\textbf{Method} & \textbf{Lift} \\
\hline
Naive Bayes & 3.27 \\
IBM1 \cite{liu2011simple} & 3.55 \\
{\bf EmTaggeR (Model 1, MuC)} & {\bf 4.38} \\
TopicWA \cite{huang2012automatic} & 4.71 \\
{\it EmTaggeR (Model 1, ALOC)} & {\bf 5.64} \\
TTM \cite{ding2013learning} & 5.87 \\
CNN+Attention-5 \cite{gong2016hashtag} & 6.42 \\
{\bf EmTaggeR (Model 2, MuC)} & {\bf 6.53} \\
{\it EmTaggeR (Model 2, ALOC)} & {\bf 7.48} \\
\hline
\end{tabular}
\end{center}
\caption{Lifts over the LDA baseline, for the different methods}
\label{tab:lift}
\vspace{-0.3in}
\end{table}

\subsection{Dataset Description}
\label{subsec:datadesc}

We collected $10\%$ of the entire Twitter data that was posted on $31^{st}$ January $2017$, using DecaHose\footnote{https://gnip.com/realtime/decahose/}.
Further, as described in Section~\ref{sec:algo}, we retain only the English tweets, containing at least one hashtag, eliminate non-ASCII characters, and remove retweets and quoted tweets to retain only tweets with original content and original hashtags.
We also remove the tweets that comprise of only hashtags that are stopwords\footnote{Appears in https://nlp.stanford.edu/IR-book/html/htmledition/dropping-common-terms-stop-words-1.html if the hash is removed from the tag}; however, if other hashtags are also present in the tweet, we retain it.
In order to ensure sufficient data volume while avoiding overtly common tweets ({\it e.g.}, {\it \#coke}), we empirically retain all the tweets containing at least one hashtag appearing in the range of $200$-$500$ times in the input dataset.
We set aside $70\%$ of the data for training, $10\%$ for validation, and $20\%$ for the final testing.
The dataset details are presented in Table~\ref{tab:datadesc}.

\subsection{Software/Tools Used}
\label{subsec:tools}

We have used the following tools, for the experiments.
(a) For stemming, we make use of the widely used Porter's Snowball Stemmer 2.0 \cite{porter2001snowball}.
(b) For tweet parsing and tokenization, we have used the Python NLTK toolkit\footnote{http://www.nltk.org}.
(c) For embedding train-test, we used Gensim \cite{rehurek2010software}.

\subsection{Hyperparameter Tuning}
\label{subsec:hyperparam}

Specifically, we fix three hyperparameters.

\begin{enumerate}
\item Setting minimum word occurrence frequency: We include only the words that occur more than a threshold number of times in the document, to avoid including insignificant words. We empirically set the minimum frequency threshold as $3$, and ignore all the words that appear less than these many times in the document.
\item Skip-size: Since we use the skip-gram algorithm of embedding (and do not use the bag-of-words model at all), we need to decide the maximum skip window size. We empirically choose to keep a window size of $4$, that is, we have skip-$4$ grams.
\item Vector dimensionality determination: We observe by studying several models that use embedding in the literature, that, typically Word2Vec embeddings between size $50$-$600$ tend to be widely used and most effective. We thus chose to perform our experiments with vector sizes of $50$, $100$, $200$, $400$ and $600$. Since, our experiments obtained the best results for vector-sizes of $50$, which is the lowest in the range, we also choose to perform an additional round of experiment with a vector size of $25$, to avoid missing out a possibly better result. As shown subsequently in Table~\ref{tab:modelperformances}, EmTaggeR consistently performs the best with vectors of size $50$ across both the models and both the performance measurement frameworks (described below).
\end{enumerate}

\subsection{Results}
\label{subsec:results}

We measure performance using two different metrics.
\begin{enumerate}
\item {\it At-Least-One Correct (ALOC):} In the ALOC model, we recommend one hashtag for each given test tweet. We consider the recommendation to be successful if the ground truth data indicates that the recommended hashtag was indeed a part of the tweet (before the hashtag was removed for the purpose of testing).
\item {\it Multiple Correct (MuC):} In the MuC model, we sort the hashtags using their embedding score. We then investigate the number of hashtags that appear in the tweet in the ground truth data, namely $K$, and pick the top-$K$ hashtags from the sorted list. Note that, by the very nature of this process of measurement, accuracy and F1-score measures here will be equivalent.
\end{enumerate}

We present the details of our experimental results below.

\subsubsection{Performance of The Different Models}
\label{subsec:different}

Table~\ref{tab:modelperformances} presents the performance for the two models used by our system, observed for both the {\it At-Least-One Correct} as well as the {\it Multiple Correct} performance metrics, effectively producing four different results.
Three factors here are noteworthy.
\begin{itemize}
\item First, the performances are consistently better for Model 1 as compared to Model 2, across both the evaluation metrics (ALOC and MuC).
\item Second, the performances peak at vector sizes $50$, as observed in our experiments that covers the range from $25$ at the low end to $600$ at the high end.
\item Third, the performances are always higher with ALOC over MuC.
\end{itemize}

\subsubsection{Establishing an LDA Baseline}
\label{subsec:baseline}

In absence of benchmark datasets for comparison, we create a LDA-based baseline score.
For this, we pick the top $3$ LDA-based topics that the test tweet is aligned to.
For each topic, we pick the one representative tweet that has the highest likelihood of belonging to that topic (amongst all the tweets that represent the topic).
We use the $3$ training tweets selected over the $3$ topics, to perform hashtag assignment to the test topic.
The LDA baseline yields F1-score of $7.79\%$.
This baseline is used subsequently to benchmark the performance of EmTaggeR (our system) with respect to the literature.

\subsubsection{Comparison with The Literature}
\label{subsec:complit}

Table~\ref{tab:lift} shows the performance of the different models under the EmTaggeR framework.
Since our system yields best-case F1 performances of \textbf{58.29\%} and \textbf{50.83\%} respectively for ALOC and MuC based measurements, the lift we obtain over the LDA baseline is $58.29/7.79$ $\approx$ \textbf{7.48} and $50.83/7.79$ $\approx$ \textbf{6.53} times.
Both of these are higher than the state-of-the-art, where the lift is 6.42 times.
Note that, both the performances reported here are obtained with Model 2.
The performances of Model 1 leave much to be desired, with ALOC and MuC yielding lifts of 5.64 and 4.38 respectively (F1-score performances of 43.95\% and 34.11\% respectively), significantly lesser than the literature.

\section{Conclusion}
\label{sec:concl}

In this work, we proposed EmTaggeR, a novel hashtag recommendation framework for Twitter.
We develop a training-testing model centered around the concept of embedding.
In the training phase, we learned word vectors associated with each hashtag, which in turn was used to derive a word embedding for each hashtag.
The testing phase constituted of computing the average word embedding of the given test post, and finding the similarity of this embedding with the known embeddings of the hashtags.
The tweets that contain the most-similar hashtag were extracted.
All the hashtags that appear in these tweets were ranked using their hashtag embedding score, and the top-$K$ hashtags were recommended.
We empirically demonstrated the effectiveness of our system, by recommending the top-$K$ ``most likely'' (most similar) hashtags for each given tweet, where $K$ is user-given number specifying the number of hashtags that a given tweet will have.
Our work provides a lift of 7.48 and 6.53 times for recommending a single hashtag and multiple hashtags to a given tweet respectively, outperforming the literature.

\bibliographystyle{abbrv}
\bibliography{bib}

\end{document}